\documentclass[10pt]{article} 
\usepackage[preprint]{tmlr}
\usepackage{graphicx}
\usepackage{xcolor}
\usepackage{tikz}
\usepackage{fp}
\usepackage{etoolbox}
\usepackage{multirow}
\usepackage{booktabs}
\usepackage{rotating}
\usepackage{subcaption}
\usepackage{placeins}
\usepackage{float}

\usepackage{amsmath,amsfonts,bm}

\def\eqref#1{equation~\ref{#1}}

\def\1{\bm{1}}

\DeclareMathAlphabet{\mathsfit}{\encodingdefault}{\sfdefault}{m}{sl}
\SetMathAlphabet{\mathsfit}{bold}{\encodingdefault}{\sfdefault}{bx}{n}

\usepackage{hyperref}
\usepackage{url}

\usepackage[dvipsnames,svgnames,x11names]{xcolor}
\usepackage{tcolorbox}
\usepackage{amsmath}
\usepackage{enumitem}
\usepackage{pifont}

\tcbuselibrary{skins,breakable}

\definecolor{questionbg}{HTML}{E8F4FD}
\definecolor{questionframe}{HTML}{2196F3}
\definecolor{englishbg}{HTML}{E8F5E9}
\definecolor{englishframe}{HTML}{4CAF50}
\definecolor{mixedbg}{HTML}{FFF3E0}
\definecolor{mixedframe}{HTML}{FF9800}
\definecolor{translationbg}{HTML}{F3E5F5}
\definecolor{translationframe}{HTML}{9C27B0}
\definecolor{correctbg}{HTML}{C8E6C9}
\definecolor{incorrectbg}{HTML}{FFCDD2}
\definecolor{highlightcn}{HTML}{FFD54F}

\newtcolorbox{questionbox}{
    enhanced,
    colback=questionbg,
    colframe=questionframe,
    fonttitle=\bfseries\large,
    title={\textcolor{white}{Input Question}},
    attach boxed title to top left={yshift=-2mm, xshift=5mm},
    boxed title style={colback=questionframe, sharp corners},
    sharp corners,
    boxrule=1.5pt,
    left=8pt, right=8pt, top=8pt, bottom=8pt,
    breakable
}

\newtcolorbox{englishbox}{
    enhanced,
    colback=englishbg,
    colframe=englishframe,
    fonttitle=\bfseries\large,
    title={\textcolor{white}{Response A: Monolingual (English)}},
    attach boxed title to top left={yshift=-2mm, xshift=5mm},
    boxed title style={colback=englishframe, sharp corners},
    sharp corners,
    boxrule=1.5pt,
    left=8pt, right=8pt, top=8pt, bottom=8pt,
    breakable
}

\newtcolorbox{mixedbox}{
    enhanced,
    colback=mixedbg,
    colframe=mixedframe,
    fonttitle=\bfseries\large,
    title={\textcolor{white}{Response B: Code-Switched (Chinese + English)}},
    attach boxed title to top left={yshift=-2mm, xshift=5mm},
    boxed title style={colback=mixedframe, sharp corners},
    sharp corners,
    boxrule=1.5pt,
    left=8pt, right=8pt, top=8pt, bottom=8pt,
    breakable
}

\newtcolorbox{translationbox}{
    enhanced,
    colback=translationbg,
    colframe=translationframe,
    fonttitle=\bfseries\large,
    title={\textcolor{white}{Translation of Code-Switched Response}},
    attach boxed title to top left={yshift=-2mm, xshift=5mm},
    boxed title style={colback=translationframe, sharp corners},
    sharp corners,
    boxrule=1.5pt,
    left=8pt, right=8pt, top=8pt, bottom=8pt,
    breakable
}

\newtcolorbox{resultbox}[1][correctbg]{
    enhanced,
    colback=#1,
    colframe=black!50,
    sharp corners,
    boxrule=1pt,
    left=5pt, right=5pt, top=5pt, bottom=5pt,
}

\title{\nc: Counterfactual Consensus\\ via Latent Space Reasoning}

\author{\name Michael Jerge \email mj6ux@virginia.edu \\
      \addr University of Virginia\\
      \AND
      \name David Evans \email evans@virginia.edu \\
      \addr University of Virginia}

\newcommand{\shortsection}[1]{\vspace*{1ex}\noindent{\bf #1.}}

\newcommand{\nc}{\textsc{NoisyCoconut}} 
\newcommand{\coconut}{\textsc{Coconut}} 

\def\month{02}
\def\year{2026}
\def\openreview{\url{https://openreview.net/forum?id=XXXX}}

\begin{document}

\maketitle

\begin{abstract}

This paper presents \nc, a novel inference-time method that enhances large language model (LLM) reliability by manipulating internal representations. Unlike fine-tuning methods that require extensive retraining, \nc\ operates directly on model representations during inference and requires no retraining. 
Rather than training models to reason in latent space, we inject controlled noise into latent trajectories to generate diverse reasoning paths. Agreement among these paths provides a confidence signal, enabling models to abstain when uncertain. We demonstrate that this approach achieves effective coverage-accuracy tradeoffs across multiple reasoning benchmarks without requiring access to training data or modification of model parameters. 
This approach provides a practical pathway to improving the reliability of LLM outputs while maintaining compatibility with existing models. Our experiments show that unanimous agreement among noise-perturbed paths reduces error rates from 40--70\% to below 15\%, enabling models to exceed 95\% accuracy on mathematical reasoning tasks through selective abstention.

\end{abstract}

\section{Introduction}
\label{sec:introduction}
Large language models (LLMs) continue to demonstrate remarkable capabilities and are increasingly deployed in high-stakes environments, including healthcare, financial services, and legal practice~\citep{Haltaufderheide_2024}. These nondeterministic models generate text autoregressively, producing one token at a time conditioned on the preceding token sequence based on a probability distribution. While this process yields fluent and coherent outputs, it provides no inherent mechanism for the model to signal uncertainty about its own generations. As a result, LLMs can produce seemingly confident but erroneous outputs, a phenomenon known as hallucination~\citep{xu2025hallucinationinevitableinnatelimitation, farquhar2024detecting}. Such errors pose particular risks in regulated sectors, where legal and regulatory obligations demand accuracy and auditability.

One approach to mitigating the risk of unreliable outputs in predictive machine learning systems is selective prediction, which allows models to abstain from giving predictions when it is likely to be incorrect~\citep{geifman2017selective, el2010foundations}. This method introduces a tradeoff between coverage (the fraction of queries answered) and accuracy (the correctness of answered queries). In classification settings, selective prediction is well-established based on softmax confidence providing a natural abstention signal~\citep{geifman2017selective, guo2017calibration}. Enabling confidence-based abstention for generative LLMs is less straightforward, however, because token-level probabilities often reflect linguistic uncertainty rather than factual correctness~\citep{kuhn2023semantic}. The main problem is how to identify instances where abstention is warranted. 

\begin{figure}[H]
    \centering
    \includegraphics[width=\textwidth]{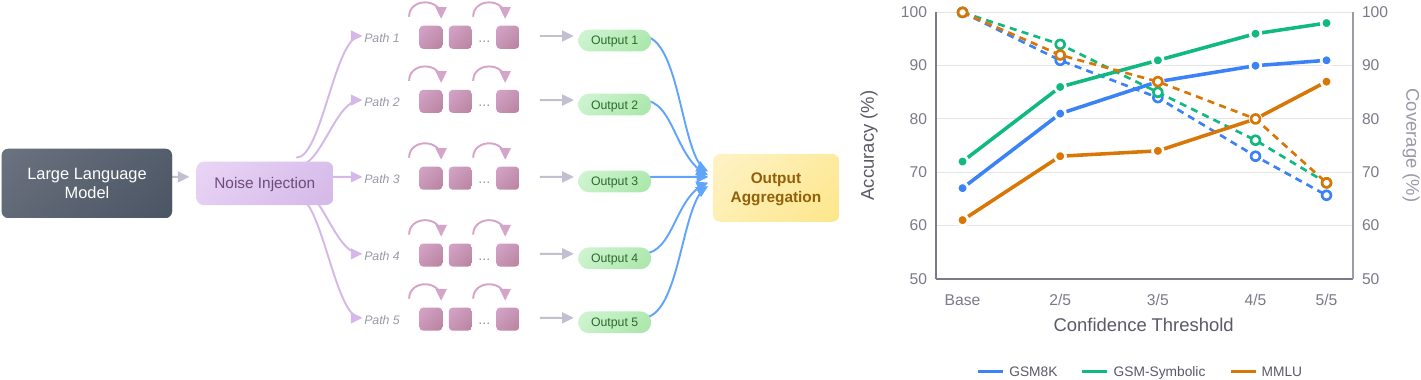}
    \caption{(Left) \nc\ architecture with noise-induced branching for diverse reasoning paths. (Right) Accuracy-coverage tradeoff for Qwen2.5-7B-Instruct across three benchmarks. Solid lines show accuracy, while the dashed lines show coverage.}
    \label{fig:coconut_branching}
\end{figure}

Several methods have been proposed for estimating the confidence of generative LLMs at inference-time. Verbalized confidence methods prompt models to explicitly rate their certainty, though such self-assessments can be poorly calibrated~\citep{kadavath2022language, xiong2024can}. Self-correction approaches~\citep[e.g.,][]{madaan2023selfrefine, shinn2023reflexion} iteratively refine outputs through self-generated feedback, though recent work suggests that intrinsic self-correction struggles for reasoning tasks without external signals~\citep{huang2024large}. Sampling-based methods generate multiple outputs and measure their consistency, using majority voting across multiple, generated paths or clustering semantically equivalent responses to compute uncertainty over meanings \citep[e.g.,][]{wang2023selfconsistency, kuhn2023semantic}. Thus, when sampled outputs disagree, the system can abstain rather than producing an unreliable answer, enabling coverage--accuracy tradeoffs. However, because these methods rely on standard autoregressive sampling, they are restricted to the diversity present in the surface-level token distribution at the output layer~\citep{holtzman2020curiouscaseneuraltext}. Consequently, they often fail to explore a sufficiently broad set of reasoning trajectories, as the sampling process is confined by the language space.

A separate line of research has begun exploring methods that operate directly on the continuous hidden states of language models, rather than through token generation~\citep{zhu2025surveylatentreasoning, goyal2023think, hao2024training}. These approaches reflect a growing recognition that the discrete, token-by-token generation process may be suboptimal for complex reasoning tasks and that reasoning in neural networks involves operations in a continuous representational space not readily accessible through token manipulations. However, latent-space methods face the same reliability challenges as their token-space counterparts, as models can produce incorrect outputs with no inherent mechanism for assessing confidence. Moreover, all of these methods require specialized architectures and training to enable latent-space reasoning.

We address this gap by extending agreement-based confidence estimation to latent space. We build on the Continuous Chain-of-Thought (\coconut) framework from \citet{hao2024training}, which trains language models to perform iterative computation in hidden states by feeding the last hidden state back as the next input embedding. Our goal differs from \coconut's---rather than training models to reason in latent space, we leverage latent representations to assess confidence in model outputs. Specifically, we introduce \nc, illustrated in Figure~\ref{fig:coconut_branching}. \nc\ injects controlled noise into latent trajectories to generate counterfactual reasoning paths. This enables selective prediction for latent-space methods, achieving effective coverage--accuracy tradeoffs without additional training.

\shortsection{Contributions} We make these main contributions: (1) We introduce \textsc{NoisyCoconut}, a training-free inference-time method that generates diverse reasoning paths through controlled noise injection in continuous latent space, enabling agreement-based confidence estimation directly on hidden states (Section~\ref{sec:noisycoconut}). (2) We demonstrate that path agreement provides a reliable proxy for prediction confidence (Section~\ref{sec:eval}). (3) We characterize the noise-accuracy relationship in latent space (Section~\ref{ssec:noise_validation}). (4) We show that \nc\ achieves effective coverage-accuracy tradeoffs across three reasoning benchmarks and five language models (Section~\ref{ssec:main_results}).

\section{Background and Related Work}
\label{sec:related_work}

Our work connects two research threads: inference-time methods for improving LLM reliability and latent-space computation. We review each area before describing the \coconut\ framework we build upon.

\subsection{Confidence Estimation and Selective Prediction}

Selective prediction for LLMs in generative settings presents fundamentally different challenges than selective prediction for classification tasks~\citep{xin2021early}. In classification, models select from a fixed set of discrete labels, but autoregressive generation produces outputs token-by-token over many steps, with an effectively unbounded space of possible responses~\citep{ren2022selective}. This distinction has motivated a substantial body of work developing uncertainty quantification methods tailored for generative models~\citep{si2022prompting, varshney2022stitch, kadavath2022language, kuhn2023semantic}. Prior works have examined selective prediction in settings where questions themselves are ambiguous, requiring models to recognize when clarification is needed rather than producing potentially incorrect answers~\citep{cole2023selectively, yin2023large}. Another line of research employs auxiliary models trained to discriminate between correct and incorrect predictions from a primary QA system~\citep{varshney2023post}, though such approaches introduce additional training requirements and may not generalize across domains.

These challenges have motivated sampling-based approaches that operationalize selective 
prediction by generating multiple outputs and using their agreement as a confidence signal 
for abstention decisions. Self-consistency~\citep{wang2023selfconsistency} samples $N$ reasoning chains from the model, then selects the most frequent final answer via majority voting. The key insight is that correct answers tend to be reachable via multiple reasoning paths, while incorrect answers arise from idiosyncratic errors unlikely to recur across samples. This approach substantially improves accuracy on arithmetic and commonsense reasoning benchmarks, though it requires generating $N$ complete token sequences per query. 

An alternative direction uses training to improve selective prediction. ASPIRE~\citep{chen2023adaptation} fine-tunes LLMs via parameter-efficient tuning to not only answer questions but also evaluate whether their generated answers are correct, producing explicit confidence scores. While effective, such approaches require task-specific training data and model access, limiting applicability.

Semantic entropy~\citep{kuhn2023semantic, farquhar2024detecting} addresses a limitation of token-level uncertainty measures since different surface forms can express the same meaning. Rather than computing entropy over token sequences, semantic entropy clusters sampled responses by meaning using bidirectional entailment, then computes entropy over these semantic equivalence classes. This provides uncertainty estimates that are invariant to paraphrasing and has been shown to detect hallucinations more reliably than token probabilities alone.

Verbalized confidence methods take a different approach, prompting models to directly express uncertainty. \citet{kadavath2022language} introduced P(True), which asks models to evaluate whether their own sampled answers are correct; the probability assigned to the token ``True'' serves as a confidence estimate. While effective for well-calibrated models, subsequent work has shown that verbalized confidence can be poorly calibrated, particularly for instruction-tuned models~\citep{xiong2024can, tian2023just}. Probing-based methods instead train classifiers on internal representations to predict correctness~\citep{azaria2023internalstatellmknows}, though these require labeled data for each task.

Recent work has extended these approaches in various directions. Confidence-Informed Self-Con\-sis\-ten\-cy~\citep{taubenfeld2025confidence} weights the majority vote by model confidence, reducing the number of samples needed. Kernel Language Entropy~\citep{nikitin2024kernel} generalizes semantic entropy using kernel methods for finer-grained uncertainty estimates.

\subsection{Inference-Time Approaches to Improving LLM Reliability}

\shortsection{Token-Space Methods}
Beyond confidence estimation, several inference-time methods seek to improve LLM reliability by structuring the generation process itself. Chain-of-Thought~\citep{wei2022chain} improves accuracy on complex tasks by eliciting intermediate steps, reducing errors that arise from single-step generation. This approach has been extended through decomposition strategies~\citep{khot2022decomposed, zhou2022least} and explicit search methods such as Tree-of-Thoughts~\citep{yao2023tree}, which explores multiple solution paths and enables backtracking.

Self-correction approaches attempt to improve reliability through iterative refinement. Self-Refine~\citep{madaan2023selfrefine} prompts models to critique and revise their own outputs, while Reflexion~\citep{shinn2023reflexion} maintains memory of past errors to guide future attempts. However, \citet{huang2024large} demonstrate that intrinsic self-correction, without external feedback, often fails to improve accuracy on reasoning tasks.

These methods operate entirely in the language and token-space, generating explicit text at each step. This limits the model to express all intermediate computation through natural language. As \citet{madaan2022text} observe, most tokens in a reasoning chain serve textual coherence rather than substantive computation.

\shortsection{Latent-Space Computation}
An alternative approach performs computation directly in the continuous hidden states of language models, bypassing token generation. Studies have shown that intermediate reasoning steps can be recovered from hidden representations~\citep{yang2024llms} and that models may employ latent processes distinct from their generated text~\citep{turpin2024language}. \citet{zhu2025surveylatentreasoning} categorize latent-space methods into activation-based approaches that expand computational depth through repeated layer processing, hidden state-based methods that maintain compressed memory states, and diffusion-based approaches enabling iterative refinement.

\shortsection{Chain of Continuous Thought (\coconut)}
Prior work has explored augmenting LLMs with special tokens that enable additional hidden-state computation, such as learnable pause tokens~\citep{goyal2023think} and filler tokens~\citep{pfau2024let}. However, these approaches require training and still operate through the token interface. \citet{hao2024training} introduced \coconut\ (Chain of Continuous Thought), which enables reasoning entirely in continuous latent space. Rather than generating intermediate tokens, \coconut\ feeds the model's last hidden state back as the next input embedding. Formally, given an initial hidden state $\mathbf{h}_0$, the model iteratively computes
$\mathbf{h}_{t+1} = f_\theta(\mathbf{h}_t)$
where $f_\theta: \mathbb{R}^d \to \mathbb{R}^d$ represents a forward pass through the transformer. This formulation enables reasoning without intermediate tokenization, achieving strong performance on tasks requiring search and planning.

Analysis of \coconut\ reveals structured latent dynamics: hidden states exhibit exploration phases with rapid movement through representation space, followed by convergence phases as solutions crystallize. The model learns to maintain stable representations of problem constraints while iteratively refining candidate solutions. Several methods have built on this foundation, including theoretical frameworks for understanding reasoning through superposition of computational states~\citep{zhu2025reasoningsuperpositiontheoreticalperspective} and demonstrations of parallel exploration in continuous representations~\citep{gozeten2025continuouschainthoughtenables}.

\shortsection{Our work bridges confidence estimation and latent-space reasoning} We observe that \coconut's continuous states provide a natural substrate for diversification, and injecting controlled noise into the hidden state trajectory induces branching into alternative reasoning paths. By measuring agreement among these paths, analogous to how self-consistency measures agreement across sampled token sequences, we obtain confidence estimates that enable coverage-accuracy tradeoffs. Unlike \coconut, which trains models for latent reasoning, \nc\ operates at inference time to assess confidence in model outputs, requiring no additional training.

\section{NoisyCoconut}
\label{sec:noisycoconut}

The core idea of \nc\ is to perturb the hidden states during the continuous hidden state space process, creating multiple reasoning trajectories that explore different regions of the solution space. Agreement among diverse reasoning paths provides stronger evidence for correctness, and allows for accuracy--coverage tradeoffs to improve reliability.

\subsection{Noise-Based Path Exploration}

The method of \nc\ is simple---we inject noise into the internal process of an LLM and aggregate results across multiple paths with different noise injection.  We sample a random noise pattern from a Gaussian distribution and inject it into the last hidden layer of the first forward pass of an LLM to create multiple reasoning paths from a common initial state. Ideally, the noise should be large enough to induce meaningful divergences, yet constrained enough to maintain coherent sequences. We hypothesize that this perturbation creates divergences in the latent space, yielding disparate reasoning chains. If these chains reach the similar conclusions, this increases our confidence that the model's prediction is correct.

Let $\mathcal{M}$ be a pre-trained language model with hidden dimension $d$. For input query $q \in \mathcal{Q}$, let $\mathbf{h}_0 = \phi(q) \in \mathbb{R}^d$ denote the initial hidden state from the first forward pass. The \nc\ process evolves as:

\begin{equation}
\mathbf{h}_{t+1} = f_\theta(\mathbf{h}_t + \boldsymbol{\eta}_t), \quad \boldsymbol{\eta}_t \sim \mathcal{N}(\mathbf{0}, \sigma_t^2 \mathbf{I}_d)
\end{equation}

where $\{\boldsymbol{\eta}_t\}$ are independent and the noise scale decays exponentially:

\begin{equation}
\sigma_t = \sigma_0 e^{-\lambda t}, \quad \sigma_0, \lambda > 0
\end{equation}

We adapt the noise based on trajectory properties. Let $\mu_t$ denote the exponentially-weighted moving average of the hidden state norm, updated as:

\begin{equation}
\mu_t = \alpha \mu_{t-1} + (1-\alpha) \|\mathbf{h}_t\|_2, \quad \alpha \in (0, 1)
\end{equation}

Thus, the adaptive noise scale is:

\begin{equation}
\sigma_t = \sigma_0 e^{-\lambda t} \cdot \frac{\|\mathbf{h}_t\|_2}{\mu_t}
\label{eq:adaptive_noise}
\end{equation}

\subsection{Path Diversity}

To ensure effective exploration of the solution space, we require sequences that are sufficiently disparate. Our goal is to produce perturbations that result in distinct paths rather than minor variations of the same solution strategy. 

For $K$ paths $\{\mathbf{h}^{(i)}_{0:T}\}_{i=1}^K$, we define pairwise trajectory diversity as:
\begin{equation}
\mathcal{D}_K = \frac{2}{K(K-1)} \sum_{1 \leq i < j \leq K} \frac{1}{T} \sum_{t=0}^{T-1} \|\mathbf{h}_t^{(i)} - \mathbf{h}_t^{(j)}\|_2
\end{equation}

The expected diversity scales with noise: $\mathbb{E}[\mathcal{D}_K] = \Omega(\sigma_0 \sqrt{dT})$ under independent path divergence.

\subsection{Aggregating Outputs}\label{ssec:aggregation}

Each of the $K$ generated paths yields an output, which we denote as $\{y^{(i)}\}_{i=1}^K \in \mathcal{Y}$. To produce the consensus output, we aggregate these generated outputs to either produce a consensus output or abstain (denoted as $\bot$). We choose a majority voting strategy by selecting the output that appears most frequently among the $K$ generated paths, treating each path equally regardless of its generation process or characteristics:

\begin{equation}
\label{eq:majority_voting}
\hat{y} = \begin{cases} 
\hat{y}^\star & \text{if } \displaystyle \exists \hat{y}^\star \in \mathcal{Y} \text{ s.t. } \sum_{i=1}^K \mathbb{I}\{y^{(i)} = \hat{y}^\star\} > \frac{K}{2} \\
\bot & \text{otherwise}
\end{cases}
\end{equation}

While Equation~\ref{eq:majority_voting} defines the general selection criteria for any $K$, for our experimental analysis (where $K=5$), we specifically categorize the consensus patterns into mutually exclusive outcomes to diagnose reliability based on the minimum number of paths that must agree for the system to produce an answer:

\begin{itemize}
    \item \textbf{Unanimous ($5/5$):} All five paths converge to the exact same answer.
    \item \textbf{Strong Majority ($4/5$):} Four paths agree on an answer, while one diverges.
    \item \textbf{Moderate Majority ($3/5$):} Three paths agree on an answer.
    \item \textbf{Minimal Plurality ($2/5$):} Two paths agree on an answer, while the other three are distinct (e.g., A, A, B, C, D).
\end{itemize}

These categories allow us to map the latent stability of the model to the correctness of the final output in our experiments, and would provide an accuracy--coverage tradeoff in a deployed system. In cases where no answer appears on more than two paths, the system will always abstain.\footnote{Although one can imagine future work that attempts to find the most likely answer even with no explicit agreement or that derives more information from the different paths by considering token probabilities or other factors, we do not explore that in this work, but keep things simple by just counting outcomes.}

\section{Evaluation and Results}
\label{sec:eval}

To understand the coverage--accuracy tradeoffs enabled by \nc, we conduct a systematic evaluation across five language models and three benchmarks. 
\autoref{ssec:setup} describes our experimental setup, including model selection, benchmarks, and implementation details. We then validate a core assumption of our method by characterizing the noise-accuracy relationship as the noise scale varies (\autoref{ssec:noise_validation}). \autoref{ssec:main_results} presents our main results, analyzing how agreement patterns among diverse reasoning paths correlate with accuracy. Our findings show that path agreement in latent space provides a strong signal for prediction reliability. Unanimous agreement among five noise-perturbed reasoning paths reduces error rates from 40--70\% to below 15\%, enabling models to achieve over 95\% accuracy on mathematical reasoning benchmarks when selectively abstaining on low-confidence predictions.

\subsection{Experimental Setup}\label{ssec:setup}

We selected representative open-source language models of similar parameter scales to evaluate the effectiveness of \nc\ across different architectures and training paradigms. The selection includes a mix of instruction-tuned models, base models, and a specialized distilled reasoning model to assess performance across different model optimization approaches. 

\begin{table}[tbh]
\centering
\caption{Language models used in evaluation}
\label{tab:models}
\begin{tabular}{lcll}
\toprule
\multicolumn{1}{c}{\textbf{Model}} & \multicolumn{1}{c}{\textbf{Parameters}} & \multicolumn{1}{c}{\textbf{Type}} \\
\midrule
Qwen2.5-7B-Instruct & 7B & Instruction-tuned \\
Llama-3.1-8B-Instruct & 8B & Instruction-tuned \\
Mixtral-8B-Instruct-v0.1 & 8B & Instruction-tuned \\
gpt-oss-20B & 20B & Foundation model \\
DeepSeek-R1-Distill-Qwen-7B & 7B & Distilled reasoning \\
\bottomrule
\end{tabular}
\end{table}

\autoref{tab:models} summarizes the models tested. Qwen2.5-7B-Instruct \citep{qwen2.5} is a recent instruction-tuned model from the Qwen family, known for strong reasoning capabilities despite its moderate size. Llama-3.1-8B-Instruct \citep{grattafiori2024llama3herdmodels} is one of Meta's smaller instruction-tuned variants in the Llama series, widely used as a foundation for many applications. Mixtral-8B-Instruct-v0.1 \citep{jiang2024mixtralexperts} provides another instruction-tuned variant, optimized specifically for following complex instructions with enhanced reasoning capabilities. gpt-oss-20B \citep{openai2025gptoss120bgptoss20bmodel} is an open-weights model that offers strong general‑purpose language abilities. Finally, DeepSeek-R1-Distill-Qwen-7B \citep{deepseekai2025deepseekr1incentivizingreasoningcapability} is a specialized distilled reasoning model that combines elements from both DeepSeek and Qwen architectures, where knowledge distillation techniques were specifically employed to enhance reasoning capabilities. 

While our \nc\ implementation works directly with standard architectures (Qwen-2.5-7B-Instruct, DeepSeek-R1-Distill-Qwen-7B), we observed that gpt-oss-20B produced degenerate outputs when using the default final-layer hidden state feedback mechanism, regardless of noise level. Thus, we use $\mathbf{h}^{(1)}$ (first layer) instead of $\mathbf{h}^{(L)}$ (final layer) for the continuous sequence. This modification enables functional operation, though it reduces the effective reasoning depth per latent pass. All other models use the standard final-layer configuration.

\shortsection{Benchmarks}
To evaluate both mathematical reasoning and broader knowledge capabilities, we selected three established benchmarks. GSM8K \citep{cobbe2021trainingverifierssolvemath} provides a challenging grade-school math word problem benchmark consisting of 1000 linguistically diverse problems requiring multi-step reasoning. GSM-Symbolic \citep{mirzadeh2025gsmsymbolicunderstandinglimitationsmathematical} represents a variant of GSM8K that replaces natural language descriptions with more symbolic representations of the same underlying problems, testing how well models handle different formulations of equivalent mathematical concepts. The MMLU (Massive Multitask Language Understanding) \citep{hendrycks2021measuringmassivemultitasklanguage} benchmark offers a comprehensive assessment covering 57 subjects across STEM, humanities, social sciences, and more, testing both factual knowledge and reasoning abilities. For each benchmark, we randomly sampled 1000 questions to ensure sufficient statistical power while maintaining computational feasibility.

\begin{figure}[btp]
    \centering
    \includegraphics[width=\textwidth]{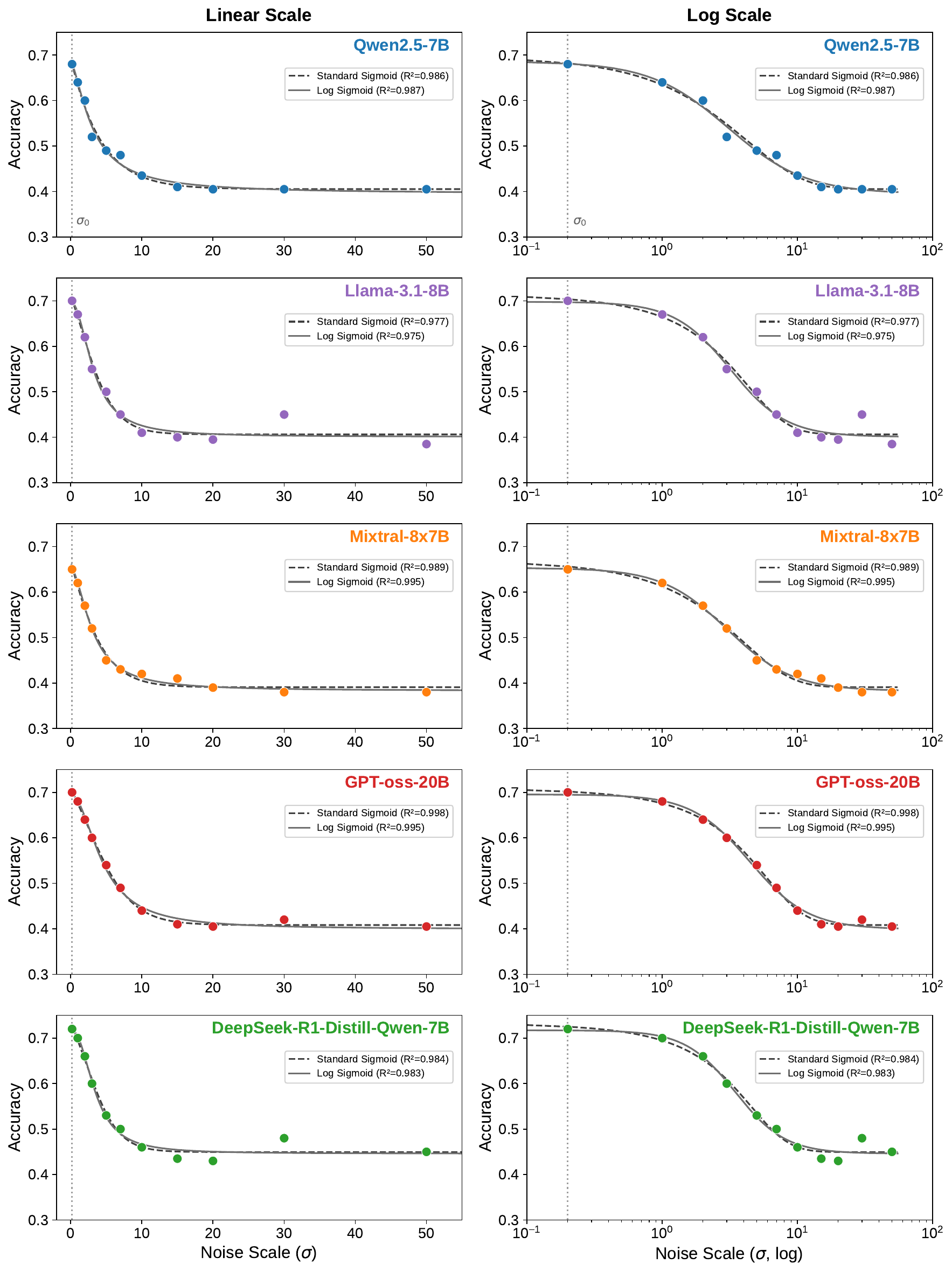}
    \caption{Accuracy degradation vs.\ noise scale $\sigma$ on linear (left) and logarithmic (right) scales. The vertical dotted line marks $\sigma_0 = 0.2$, chosen to balance perturbation strength with model performance (see Section~\ref{sec:ablation}). Sigmoid fits achieve $R^2 \geq 0.94$ for all models.}
    \label{fig:log-decay}
\end{figure}

\subsection{Validating Impact of Noise}\label{ssec:noise_validation}
A fundamental assumption of \nc\ is that injecting noise into hidden states produces controlled and predictable degradation in model performance. To validate this assumption and inform our choice of noise level, we systematically measured accuracy across noise scales $\sigma \in [0, 50]$, where $\sigma$ represents the ratio of noise norm to hidden state norm ($\|\boldsymbol{\epsilon}\|_2 = \sigma \|\mathbf{h}\|_2$).

\autoref{fig:log-decay} summarizes the results, showing a characteristic sigmoid decay pattern across all models, with fits achieving $R^2 \geq 0.94$. Accuracy remains relatively stable at low noise levels before undergoing steep degradation, eventually plateauing near random performance. This consistent pattern across architectures confirms that noise injection produces predictable, exploitable effects.

Based on these results, we selected $\sigma_0 = 0.2$ (marked by the vertical dotted line in \autoref{fig:log-decay}) as the noise level for our main experiments. This value lies in the early portion of the degradation curve, where accuracy remains within 2--5\% of unperturbed performance. This choice introduces sufficient perturbation to generate meaningful diversity in generated paths, while avoiding excessive degradation that would compromise individual path quality. Further analysis of this trade-off is provided in Appendix~\ref{sec:ablation}.

\subsection{Results and Analysis}\label{ssec:main_results}

We implemented \nc\ as described in \autoref{sec:noisycoconut}, with several key hyperparameters selected to balance performance and computational cost. 

For our main experiments, we use five reasoning paths per problem, which our ablation studies (\autoref{sec:ablation}) show provides a good efficiency-performance tradeoff.
As mentioned earlier, based on our experiments in \autoref{ssec:noise_validation} we set the noise scale to $\sigma_0 = 0.2$. 
Noise injections were performed at the first hidden state after the first forward pass. Early injection enables greater exploration of the reasoning space and aligns with our goal of generating diverse reasoning trajectories from the outset. We set a maximum of eight latent thinking steps to allow sufficient reasoning depth, following the same constraint established in the original \coconut\ work.\footnote{Note that these component-wise ablations do not capture interaction effects between hyperparameters.}

For each model and benchmark combination, we first establish a baseline by evaluating the model using standard greedy decoding without inference-time modifications. We then apply \textsc{Noisy Coconut}, generating five ($K=5$) distinct reasoning paths for each test example by injecting controlled noise into the hidden states. We analyze the generated paths using the aggregation outcomes defined in Section~\ref{sec:noisycoconut}. While we track distinct outcomes such as ``Split Votes'' for error analysis, our primary performance evaluation focuses on the cumulative confidence threshold. We perform this generation step once per test example, and then analyze the impact of applying the different aggregation functions to this static set of outputs. This allows us to observe how the consensus output $\hat{y}$, which is either a specific answer $\hat{y}^\star$ or an abstention $\perp$, changes as a function of the strictness of the agreement threshold. 

\begin{figure*}[t]
    \centering
    \subfloat[GSM8K\label{fig:gsm8k}]{%
        \includegraphics[width=0.32\textwidth]{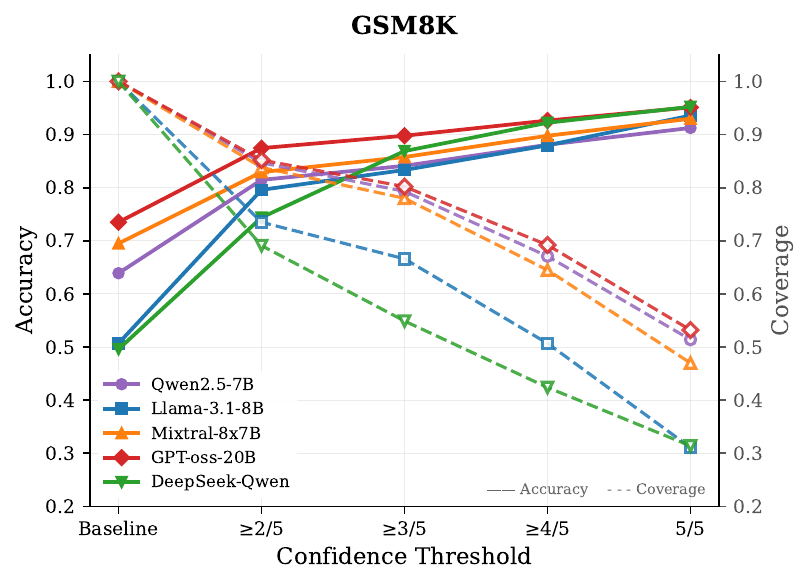}%
    }
    \hfill
    \subfloat[GSM-Symbolic\label{fig:gsm_symbolic}]{%
        \includegraphics[width=0.32\textwidth]{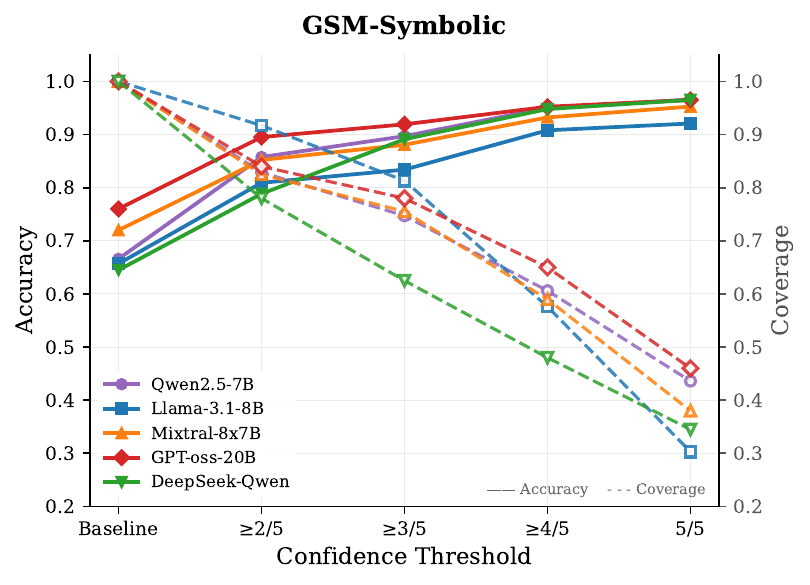}%
    }
    \hfill
    \subfloat[MMLU\label{fig:mmlu}]{%
        \includegraphics[width=0.32\textwidth]{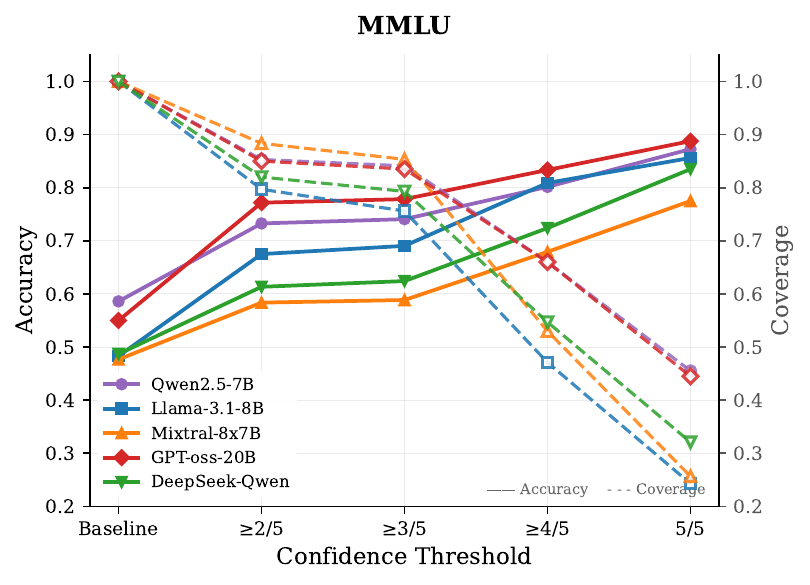}%
    }
    \caption{Coverage-accuracy trade-off across confidence thresholds. Solid lines denote accuracy and dashed lines denote coverage. As the agreement threshold increases from baseline to unanimous, accuracy improves while coverage decreases correspondingly.}
    \label{fig:coverage_accuracy}
\end{figure*}

Figure~\ref{fig:coverage_accuracy} illustrates the fundamental trade-off enabled by our method. We observe a consistent relationship between the agreement threshold and accuracy. As the threshold increases from a plurality ($\ge 2/5$ agreement) through majority requirements ($\ge 3/5$) to unanimous agreement ($\ge 5/5$), the reliability improves significantly.

Across the five language models and three benchmarks, the path agreement confidence measure enabled by \nc\ provides a trade-off between \emph{coverage} (the proportion of questions for which the system provides an answer) and \emph{accuracy} (the correctness of answers output when the model does not abstain).  

We observe a consistent relationship between agreement threshold and accuracy, with performance improving as the threshold increases from the baseline (single-path inference) through $\ge 2$, $\ge 3$, $\ge 4$ and unanimous agreement thresholds.

Our results demonstrate that \nc\ offers a mechanism for enhancing predictive accuracy through selective abstention on low-confidence predictions.
The magnitude of improvement on mathematical benchmarks is particularly noteworthy. On GSM8K, no baseline model exceeds 75\% accuracy, yet at the unanimous (5/5) agreement threshold, all models exceed 90\% accuracy, with coverage ranging from 31.1\% (Llama-3.1-8B) to 53.2\% (gpt-oss-20B). The largest accuracy improvement is observed for DeepSeek-Qwen, which improves from 62.0\% accuracy at the baseline to 95.2\% at unanimous agreement, while maintaining 31.4\% coverage.
Similar patterns emerge across all evaluated models, suggesting that path agreement constitutes an effective method for measuring prediction reliability.

\begin{figure*}[tb]
\centering

    \csdef{GSM8kQwen2T2}{17.6}
    \csdef{GSM8kQwen2T3}{20.2}
    \csdef{GSM8kQwen2T4}{24.2}
    \csdef{GSM8kQwen2T5}{27.3}

    \csdef{GSM8kLlama31T2}{28.9}
    \csdef{GSM8kLlama31T3}{32.6}
    \csdef{GSM8kLlama31T4}{37.3}
    \csdef{GSM8kLlama31T5}{42.9}

    \csdef{GSM8kMixtralT2}{13.4}
    \csdef{GSM8kMixtralT3}{16.3}
    \csdef{GSM8kMixtralT4}{20.3}
    \csdef{GSM8kMixtralT5}{23.5}

    \csdef{GSM8kGPTossT2}{13.9}
    \csdef{GSM8kGPTossT3}{16.3}
    \csdef{GSM8kGPTossT4}{19.1}
    \csdef{GSM8kGPTossT5}{21.6}

    \csdef{GSM8kDeepSeekT2}{24.8}
    \csdef{GSM8kDeepSeekT3}{37.3}
    \csdef{GSM8kDeepSeekT4}{42.6}
    \csdef{GSM8kDeepSeekT5}{45.6}

    \csdef{GSMSymbolicQwen2T2}{19.2}
    \csdef{GSMSymbolicQwen2T3}{23.1}
    \csdef{GSMSymbolicQwen2T4}{28.4}
    \csdef{GSMSymbolicQwen2T5}{30.0}

    \csdef{GSMSymbolicLlama31T2}{15.2}
    \csdef{GSMSymbolicLlama31T3}{17.7}
    \csdef{GSMSymbolicLlama31T4}{25.1}
    \csdef{GSMSymbolicLlama31T5}{26.4}

    \csdef{GSMSymbolicMixtralT2}{13.2}
    \csdef{GSMSymbolicMixtralT3}{16.1}
    \csdef{GSMSymbolicMixtralT4}{21.2}
    \csdef{GSMSymbolicMixtralT5}{23.3}

    \csdef{GSMSymbolicGPTossT2}{13.5}
    \csdef{GSMSymbolicGPTossT3}{15.9}
    \csdef{GSMSymbolicGPTossT4}{19.2}
    \csdef{GSMSymbolicGPTossT5}{20.5}

    \csdef{GSMSymbolicDeepSeekT2}{14.3}
    \csdef{GSMSymbolicDeepSeekT3}{24.6}
    \csdef{GSMSymbolicDeepSeekT4}{30.3}
    \csdef{GSMSymbolicDeepSeekT5}{32.0}

    \csdef{MMLUQwen2T2}{14.7}
    \csdef{MMLUQwen2T3}{15.5}
    \csdef{MMLUQwen2T4}{21.6}
    \csdef{MMLUQwen2T5}{28.7}

    \csdef{MMLULlama31T2}{19.2}
    \csdef{MMLULlama31T3}{20.7}
    \csdef{MMLULlama31T4}{32.6}
    \csdef{MMLULlama31T5}{37.3}

    \csdef{MMLUMixtralT2}{10.7}
    \csdef{MMLUMixtralT3}{11.2}
    \csdef{MMLUMixtralT4}{20.2}
    \csdef{MMLUMixtralT5}{29.8}

    \csdef{MMLUGPTossT2}{22.2}
    \csdef{MMLUGPTossT3}{22.8}
    \csdef{MMLUGPTossT4}{28.3}
    \csdef{MMLUGPTossT5}{33.8}

    \csdef{MMLUDeepSeekT2}{12.6}
    \csdef{MMLUDeepSeekT3}{13.7}
    \csdef{MMLUDeepSeekT4}{23.7}
    \csdef{MMLUDeepSeekT5}{34.8}

\resizebox{\linewidth}{!}{%
\begin{tikzpicture}[x=1cm, y=1cm]

    \definecolor{COLFmax}{HTML}{1a9641}
    \definecolor{COLFmid}{HTML}{ffffbf}
    
    \def\Fmax{46}
    \def\Fmin{0}

    \def\cellw{1.2}
    \def\cellh{0.8}

    \def\colstart{4.2}

    \foreach \bench [count=\benchidx from 0] in {GSM8k, GSMSymbolic, MMLU} {

        \pgfmathsetmacro{\benchx}{\colstart + 1.5*\cellw + \benchidx*5.5*\cellw}
        \node[font=\bfseries\small] at (\benchx, 4.9) {
            \ifnum\benchidx=0 GSM8K\fi
            \ifnum\benchidx=1 GSM-Symbolic\fi
            \ifnum\benchidx=2 MMLU\fi
        };

        \foreach \thresh [count=\threshidx from 0] in {{2/5}, {3/5}, {4/5}, {5/5}} {
            \pgfmathsetmacro{\tx}{\colstart + \threshidx*\cellw + \benchidx*5.5*\cellw}
            \node[font=\scriptsize, anchor=south] at (\tx, 4.1) {$\geq$\thresh};
        }

        \foreach \model [count=\modelidx from 0] in {Qwen2, Llama31, Mixtral, GPToss, DeepSeek} {

            \ifnum\benchidx=0
                \pgfmathsetmacro{\my}{4 - \modelidx*\cellh - 0.4}
                \node[font=\scriptsize, anchor=east] at (3.4, \my) {
                    \ifnum\modelidx=0 Qwen\fi
                    \ifnum\modelidx=1 Llama\fi
                    \ifnum\modelidx=2 Mixtral\fi
                    \ifnum\modelidx=3 gpt-oss\fi
                    \ifnum\modelidx=4 DeepSeek\fi
                };
            \fi

            \foreach \thresh [count=\threshidx from 0] in {T2, T3, T4, T5} {
                \pgfmathsetmacro{\cx}{\colstart + \threshidx*\cellw + \benchidx*5.5*\cellw}
                \pgfmathsetmacro{\cy}{4 - \modelidx*\cellh - 0.4}

                \pgfmathsetmacro{\myvalue}{\csuse{\bench\model\thresh}}

                \pgfmathsetmacro{\intensity}{min(100, \myvalue/\Fmax*100)}

                \colorlet{cellcolor}{COLFmax!\intensity!COLFmid}

                \fill[cellcolor, rounded corners=1pt] 
                    (\cx-\cellw/2+0.05, \cy-\cellh/2+0.05) 
                    rectangle (\cx+\cellw/2-0.05, \cy+\cellh/2-0.05);

                \draw[white, line width=0.5pt, rounded corners=1pt] 
                    (\cx-\cellw/2+0.05, \cy-\cellh/2+0.05) 
                    rectangle (\cx+\cellw/2-0.05, \cy+\cellh/2-0.05);

                \pgfmathsetmacro{\textcolor}{\intensity > 60 ? 1 : 0}
                \ifnum\textcolor=1
                    \node[font=\scriptsize\bfseries, white] at (\cx, \cy) {+\pgfmathprintnumber[fixed, precision=1]{\myvalue}};
                \else
                    \node[font=\scriptsize\bfseries, black!80] at (\cx, \cy) {+\pgfmathprintnumber[fixed, precision=1]{\myvalue}};
                \fi
            }
        }
        
    }
    
\end{tikzpicture}%
}

\caption{Cumulative accuracy improvement over baseline (in percentage points) at each confidence threshold. Higher agreement requirements yield larger improvements across all models and benchmarks. Values represent the accuracy gain when the model provides an answer at that confidence threshold.}
\label{fig:cumulative_accuracy_heatmap}
\end{figure*}

\begin{figure*}[tb]
\centering

    \csdef{GSM8kQwen2E2}{57.4}
    \csdef{GSM8kQwen2E3}{37.7}
    \csdef{GSM8kQwen2E4}{22.3}
    \csdef{GSM8kQwen2E5}{8.8}

    \csdef{GSM8kLlama31E2}{56.5}
    \csdef{GSM8kLlama31E3}{31.4}
    \csdef{GSM8kLlama31E4}{20.9}
    \csdef{GSM8kLlama31E5}{6.4}

    \csdef{GSM8kMixtralE2}{55.2}
    \csdef{GSM8kMixtralE3}{33.3}
    \csdef{GSM8kMixtralE4}{18.9}
    \csdef{GSM8kMixtralE5}{7.0}

    \csdef{GSM8kGPTossE2}{50.0}
    \csdef{GSM8kGPTossE3}{28.2}
    \csdef{GSM8kGPTossE4}{15.6}
    \csdef{GSM8kGPTossE5}{4.9}

    \csdef{GSM8kDeepSeekE2}{73.9}
    \csdef{GSM8kDeepSeekE3}{31.2}
    \csdef{GSM8kDeepSeekE4}{16.4}
    \csdef{GSM8kDeepSeekE5}{4.8}

    \csdef{GSMSymbolicQwen2E2}{50.0}
    \csdef{GSMSymbolicQwen2E3}{33.3}
    \csdef{GSMSymbolicQwen2E4}{8.8}
    \csdef{GSMSymbolicQwen2E5}{3.4}

    \csdef{GSMSymbolicLlama31E2}{38.5}
    \csdef{GSMSymbolicLlama31E3}{34.6}
    \csdef{GSMSymbolicLlama31E4}{10.6}
    \csdef{GSMSymbolicLlama31E5}{7.9}

    \csdef{GSMSymbolicMixtralE2}{45.7}
    \csdef{GSMSymbolicMixtralE3}{30.3}
    \csdef{GSMSymbolicMixtralE4}{10.5}
    \csdef{GSMSymbolicMixtralE5}{4.7}

    \csdef{GSMSymbolicGPTossE2}{41.7}
    \csdef{GSMSymbolicGPTossE3}{24.6}
    \csdef{GSMSymbolicGPTossE4}{7.9}
    \csdef{GSMSymbolicGPTossE5}{3.5}

    \csdef{GSMSymbolicDeepSeekE2}{62.6}
    \csdef{GSMSymbolicDeepSeekE3}{29.7}
    \csdef{GSMSymbolicDeepSeekE4}{9.6}
    \csdef{GSMSymbolicDeepSeekE5}{3.5}

    \csdef{MMLUQwen2E2}{83.3}
    \csdef{MMLUQwen2E3}{48.1}
    \csdef{MMLUQwen2E4}{35.6}
    \csdef{MMLUQwen2E5}{12.7}

    \csdef{MMLULlama31E2}{61.0}
    \csdef{MMLULlama31E3}{50.5}
    \csdef{MMLULlama31E4}{24.1}
    \csdef{MMLULlama31E5}{14.4}

    \csdef{MMLUMixtralE2}{55.6}
    \csdef{MMLUMixtralE3}{56.0}
    \csdef{MMLUMixtralE4}{41.1}
    \csdef{MMLUMixtralE5}{22.5}

    \csdef{MMLUGPTossE2}{60.0}
    \csdef{MMLUGPTossE3}{42.9}
    \csdef{MMLUGPTossE4}{27.9}
    \csdef{MMLUGPTossE5}{11.2}

    \csdef{MMLUDeepSeekE2}{70.4}
    \csdef{MMLUDeepSeekE3}{59.8}
    \csdef{MMLUDeepSeekE4}{43.4}
    \csdef{MMLUDeepSeekE5}{16.5}

\resizebox{\linewidth}{!}{%
\begin{tikzpicture}[x=1cm, y=1cm]

    \definecolor{COLFmax}{HTML}{d73027}
    \definecolor{COLFmid}{HTML}{fee08b}
    \definecolor{COLFmin}{HTML}{1a9850}
    
    \def\Fmax{85}
    \def\Fmin{0}

    \def\cellw{1.2}
    \def\cellh{0.8}

    \def\colstart{4.2}

    \foreach \bench [count=\benchidx from 0] in {GSM8k, GSMSymbolic, MMLU} {

        \pgfmathsetmacro{\benchx}{\colstart + 1.5*\cellw + \benchidx*5.5*\cellw}
        \node[font=\bfseries\small] at (\benchx, 4.9) {
            \ifnum\benchidx=0 GSM8K\fi
            \ifnum\benchidx=1 GSM-Symbolic\fi
            \ifnum\benchidx=2 MMLU\fi
        };

        \foreach \thresh [count=\threshidx from 0] in {{2/5}, {3/5}, {4/5}, {5/5}} {
            \pgfmathsetmacro{\tx}{\colstart + \threshidx*\cellw + \benchidx*5.5*\cellw}
            \node[font=\scriptsize, anchor=south] at (\tx, 4.1) {\thresh};
        }

        \foreach \model [count=\modelidx from 0] in {Qwen2, Llama31, Mixtral, GPToss, DeepSeek} {

            \ifnum\benchidx=0
                \pgfmathsetmacro{\my}{4 - \modelidx*\cellh - 0.4}
                \node[font=\scriptsize, anchor=east] at (3.4, \my) {
                    \ifnum\modelidx=0 Qwen\fi
                    \ifnum\modelidx=1 Llama\fi
                    \ifnum\modelidx=2 Mixtral\fi
                    \ifnum\modelidx=3 gpt-oss\fi
                    \ifnum\modelidx=4 DeepSeek\fi
                };
            \fi

            \foreach \thresh [count=\threshidx from 0] in {E2, E3, E4, E5} {
                \pgfmathsetmacro{\cx}{\colstart + \threshidx*\cellw + \benchidx*5.5*\cellw}
                \pgfmathsetmacro{\cy}{4 - \modelidx*\cellh - 0.4}

                \pgfmathsetmacro{\myvalue}{\csuse{\bench\model\thresh}}

                \pgfmathsetmacro{\intensity}{min(100, \myvalue/\Fmax*100)}

                \ifdim\myvalue pt<42.5pt

                    \pgfmathsetmacro{\localint}{\myvalue/42.5*100}
                    \colorlet{cellcolor}{COLFmid!\localint!COLFmin}
                \else

                    \pgfmathsetmacro{\localint}{(\myvalue-42.5)/42.5*100}
                    \colorlet{cellcolor}{COLFmax!\localint!COLFmid}
                \fi

                \fill[cellcolor, rounded corners=1pt] 
                    (\cx-\cellw/2+0.05, \cy-\cellh/2+0.05) 
                    rectangle (\cx+\cellw/2-0.05, \cy+\cellh/2-0.05);

                \draw[white, line width=0.5pt, rounded corners=1pt] 
                    (\cx-\cellw/2+0.05, \cy-\cellh/2+0.05) 
                    rectangle (\cx+\cellw/2-0.05, \cy+\cellh/2-0.05);

                \pgfmathsetmacro{\usewhite}{(\myvalue > 55) || (\myvalue < 15) ? 1 : 0}
                \ifnum\usewhite=1
                    \node[font=\scriptsize\bfseries, white] at (\cx, \cy) {\pgfmathprintnumber[fixed, precision=0]{\myvalue}\%};
                \else
                    \node[font=\scriptsize\bfseries, black!80] at (\cx, \cy) {\pgfmathprintnumber[fixed, precision=0]{\myvalue}\%};
                \fi
            }
        }
        
    }
    
\end{tikzpicture}%
}

\caption{Error rates at each agreement level across models and benchmarks. Lower agreement levels (2/5, 3/5) consistently exhibit higher error rates (shown in red/orange), while high-agreement predictions (4/5, 5/5) achieve substantially lower error rates (shown in green). This confirms that path disagreement effectively identifies unreliable predictions suitable for abstention.}
\label{fig:error_rate_heatmap}
\end{figure*}

\paragraph{Accuracy Improvement Analysis.}
\autoref{fig:cumulative_accuracy_heatmap} shows the accuracy gains achieved at each confidence threshold relative to baseline performance. The heatmap reveals consistent improvements across all model--benchmark combinations, with gains ranging from approximately 10 percentage points at the lowest agreement threshold to over 40 percentage points with unanimous agreement for certain models. Models with lower baseline accuracy exhibit the largest absolute improvements, achieving near-parity with stronger models when predictions are restricted to high-agreement instances. This finding suggests that \nc\ can serve as an equalizing mechanism, enabling weaker models to approach the reliability of stronger counterparts on the subset of questions where they exhibit high confidence. The continuous increase in improvement magnitude from further corroborates the effectiveness of agreement level as a confidence signal.

\paragraph{Error Distribution.}
\autoref{fig:error_rate_heatmap} presents another perspective through examination of error rates stratified by agreement level. Test examples where there is low path agreement exhibit substantially elevated error rates, typically ranging from 40\% to 70\%, whereas high-agreement predictions demonstrate markedly lower error rates, generally below 15\% for instances of unanimous agreement.  These results are remarkably consistent across the five tested models and three benchmarks.

This concentration of errors within low-agreement categories provides direct empirical validation for the assumption underlying \nc: path disagreement serves as a reliable indicator of prediction uncertainty, effectively identifying instances where model outputs are unreliable and where abstention or additional verification would be warranted. The consistency of this pattern across diverse models and benchmarks underscores the robustness of agreement-based confidence estimation as a general-purpose mechanism for uncertainty quantification in language model inference.

\section{Discussion}
\label{sec:discussion}

By injecting controlled noise into hidden states rather than operating at the token level, \nc\ provides an inference-time method that introduces an accuracy--coverage tradeoff which can be used to enhance LLM reliability without any need for retraining.  The consistent relationship between path agreement and accuracy suggests that uncertainty in LLMs may be visible in latent space. When noise perturbations lead to divergent reasoning paths, this likely indicates the model is operating in a region of representational instability where small changes in hidden states lead to substantially different outputs. Conversely, when perturbed paths converge to the same answer, the model appears to be in a stable attractor basin where the solution is robust to perturbation. This interpretation aligns with recent theoretical work analyzing how reasoning emerges through superposition of computational states in continuous representations \citep{zhu2025reasoning}.

The sigmoid degradation pattern observed in Figure~\ref{fig:log-decay}, with fits achieving $R^2 \geq 0.94$ across all models, suggests a phase transition in model behavior. At low noise levels, the model's reasoning process is robust enough to absorb perturbations without changing outputs. Beyond a critical threshold, perturbations overwhelm the signal, causing rapid degradation to near-random performance. The consistency of this pattern across architectures, from instruction-tuned models to distilled reasoning models, indicates this may be a fundamental property of how transformer representations encode reasoning processes rather than an artifact of specific training procedures.

Our finding that weaker baseline models exhibit larger absolute improvements under high-agreement filtering merits further investigation. One hypothesis is that weaker models possess latent capabilities that are inconsistently activated during standard inference. \textsc{NoisyCoconut} may function as a capability elicitation mechanism, identifying instances where the model ``knows'' the answer but requires favorable initialization to reliably surface it. This speculation connects to work on latent knowledge in language models \citep{christiano2021eliciting} and suggests that agreement-based filtering could complement techniques designed to extract reliable knowledge from uncertain models.

\subsection{Limitations}

\paragraph{Generalizability.}
\nc\ requires the ability to access and perturb internal model states, limiting applicability to open-weight models or settings with sufficient API access. The method cannot currently be applied to closed models accessible only through text-based APIs. As latent-space methods gain prominence, API designs that expose intermediate representations could enable broader application of techniques like ours \citep{gao2023retrieval}.

Our method's effectiveness varies across architectures. Most notably, gpt-oss-20B required using first-layer hidden states rather than final-layer states for the continuous feedback loop, as the default configuration produced degenerate outputs. This sensitivity suggests that the structure of hidden representations differs meaningfully across model families, and optimal noise injection strategies may need architecture-specific tuning. Understanding why certain architectures respond differently to latent perturbation remains an open question.

\paragraph{Discrete Responses.}
Our evaluation focused on mathematical reasoning and knowledge-intensive tasks where the set of responses is small (multiple-choice questions) and agreement is well-defined. Extending \textsc{NoisyCoconut} to open-ended generation tasks where ``agreement'' requires semantic similarity rather than exact matching poses a difficult, but we think not impenetrable, challenge.  Integration with semantic entropy methods \citep{kuhn2023semantic, farquhar2024detecting} could enable agreement-based confidence for tasks like summarization or translation, where multiple valid outputs exist. Additionally, exploring whether latent perturbation can improve factuality in long-form generation, where hallucination risks compound, represents a practically important direction.

\paragraph{Design space exploration.} While we identified $\sigma_0 = 0.2$ as effective across our experiments, the optimal noise scale likely depends on task difficulty, model capacity, and input characteristics. Our current approach uses fixed hyperparameters, but adaptive noise scaling based on input uncertainty or model confidence could improve robustness. The exponentially weighted moving average adaptation in Equation~\ref{eq:adaptive_noise} represents a first step, but more sophisticated approaches drawing on adaptive gradient methods \citep{kingma2015adam} or learned noise schedules \citep{ho2020denoising} may prove beneficial. 

Our evaluation was limited to consensus by voting as the aggregation strategy, where outputs are compared via exact match. The vast space of possible aggregation strategies remains largely unexplored. More sophisticated approaches such as learned aggregation functions that consider output semantics, confidence-weighted schemes based on token-level probabilities, or methods that leverage partial agreement structure may yield improved performance. Additionally, our exact-match criterion for agreement is well-suited to mathematical reasoning tasks with unique correct answers but may be overly restrictive for domains where semantically equivalent but lexically distinct outputs are valid. Integrating semantic similarity measures \citep{kuhn2023semantic} or embedding-based comparison into the aggregation step could extend \textsc{NoisyCoconut} to open-ended generation tasks where multiple valid phrasings exist. Making the aggregation adaptive would also offer additional opportunities, using additional executions only when necessary to reduce cost, and providing finer grained methods for determining when to abstain, when to invest more compute, and when there is sufficient confidence to produce a reliable output.

\paragraph{Cost.}
A straightforward implementation of \nc\ requires computational overhead to generating $K$ paths that  scales approximately linearly with $K$. While we believe $K = 5$ is already a reasonable cost for improved reliability in many settings, this may be prohibitively expensive for many applications where inference costs already dominate. 

Current \textsc{NoisyCoconut} generates paths independently, missing opportunities for computation sharing. Speculative decoding techniques \citep{leviathan2023fast, chen2023accelerating} demonstrate that parallel verification can be substantially cheaper than parallel generation. Analogously, architectures that share early computation across paths while branching only at key decision points could reduce overhead while maintaining diversity. Tree-structured approaches \citep{yao2023tree} provide one template, though adapting these to continuous latent space rather than discrete token space requires further development.

Techniques for early termination when paths show rapid convergence could mitigate this cost. It may also be the case that multiple executions of a smaller model that can be run locally within the \nc\ framework can be used to provide more reliable results for lower cost than would be required for a single execution of an expensive proprietary model. 

Controlling the computational budget allocated to reasoning is an active research area. Length Controlled Policy Optimization (LCPO) enables precise control over reasoning length, allowing models to generate outputs adhering to user-specified length constraints \citep{aggarwal2025l1}. Similarly, the simple test-time scaling (s1) approach introduces ``budget forcing'' as a technique to control test-time computation by either terminating the model's thinking process early or extending it to encourage further reasoning \citep{muennighoff2025s1}. Both approaches reveal that models can adapt their reasoning strategies based on available computational resources. Combining these insights with \textsc{NoisyCoconut} suggests an adaptive approach: allocate more reasoning paths to instances showing early signs of disagreement while terminating early when paths rapidly converge. Such selective computation could maintain reliability gains while substantially reducing average-case overhead.

\subsection{Future Directions}

\paragraph{Integration with Learned Latent Reasoning.}
One promising direction involves architectures specifically designed to leverage the latent space directly for extended reasoning. Recent work has introduced recurrent latent reasoning architectures that enable scaling test-time computation through iterative processing in latent space \citep{tan2025think, orlicki2025beyond, xu2025softcot}. These approaches employ core recurrent blocks that can be executed multiple times before producing a final output, allowing models to perform more computation without generating additional tokens. Since these architectures explicitly optimize representations for iterative reasoning, they may exhibit even more structured uncertainty geometry that \textsc{NoisyCoconut} could exploit. Investigating whether our agreement signal remains predictive—or becomes even more informative—when applied to such architectures represents a natural extension.

\paragraph{Language Mixing and Code-Switching Phenomena.}
During our experiments with varying noise thresholds, we observed an unexpected phenomenon: at certain perturbation levels, models exhibited spontaneous language mixing, producing reasoning traces that alternated between English and other languages (e.g., Chinese) despite receiving English-only prompts (Appendix \ref{sec:language_mixing}). Intriguingly, these code-switched outputs sometimes yielded correct answers where monolingual reasoning failed, suggesting that language mixing may activate alternative reasoning pathways or access knowledge encoded differently across linguistic subspaces. This observation aligns with recent systematic studies by \citet{wang2025language}, who demonstrate that language mixing in reasoning language models reflects latent processing preferences and that forcing models to reason in specific scripts can notably improve accuracy. The broader phenomenon connects to emerging research on latent chain-of-thought reasoning, which decouples reasoning from explicit language generation \citep{chen2025reasoning}, and work showing that activation-space perturbations can encode complex reasoning patterns \citep{pfau2024uncovering}. However, we did not pursue this direction further, as we lacked a principled mechanism to predict when code-switching would be beneficial or to reliably induce it. Developing methods to detect instances where language mixing improves reasoning—and techniques to controllably trigger such behavior—represents a promising avenue for future work, potentially combining our agreement-based confidence estimation with script-aware decoding strategies.

\paragraph{Theoretical Foundations.}
The empirical success of agreement-based confidence estimation invites theoretical investigation. Why should path diversity in latent space correlate with correctness? Recent work by \citet{zhu2025reasoning} provides one lens through their analysis of reasoning via superposition, suggesting that correct solutions may correspond to more stable superposition states. \citet{gozeten2025continuous} demonstrate that continuous representations enable parallel exploration of solution paths, which may explain why noise injection effectively diversifies reasoning. Developing a formal framework connecting latent geometry, perturbation stability, and output correctness could guide principled improvements to noise injection strategies and provide guarantees on when agreement-based confidence is reliable.

\bibliography{tmlr}
\bibliographystyle{tmlr}

\appendix

\clearpage
\section{Main Results}
\label{sec:main_results}

\begin{table}[ht]
\centering
\footnotesize
\setlength{\tabcolsep}{3pt}
\resizebox{\textwidth}{!}{%
\begin{tabular}{@{}lcccccc@{}}
\toprule
& & Question Count & Correct & No Answer & Incorrect & Accuracy (\%) \\
\midrule
\multirow{6}{*}{\rotatebox[origin=c]{90}{%
\begin{tabular}{c}\\
\scriptsize Qwen2.5-7B\\
\end{tabular}}}
& Baseline (No majority) & 1000 & 639 & 0 & 261 & 71.00 \\
\cmidrule(lr){2-7}
& Minimal Majority (2/5) & 54 & 23 & 0 & 31 & 42.59 \\
& Moderate Majority (3/5) & 122 & 76 & 0 & 46 & 62.30 \\
& Other & 31 & 2 & 0 & 29 & 6.45 \\
& Split Vote (2/2) & 22 & 12 & 0 & 10 & 54.55 \\
& Strong Majority (4/5) & 157 & 122 & 0 & 35 & 77.71 \\
& Unanimous (5/5) & 514 & 469 & 0 & 45 & 91.25 \\
\midrule
\multirow{6}{*}{\rotatebox[origin=c]{90}{%
\begin{tabular}{c}\\
\scriptsize Llama-3.1-8B\\
\end{tabular}}}
& Baseline (No majority) & 1000 & 507 & 0 & 293 & 63.38 \\
\cmidrule(lr){2-7}
& Minimal Majority (2/5) & 69 & 30 & 0 & 39 & 43.48 \\
& Moderate Majority (3/5) & 159 & 109 & 0 & 50 & 68.55 \\
& Other & 31 & 1 & 0 & 30 & 3.23 \\
& Split Vote (2/2) & 34 & 10 & 0 & 24 & 29.41 \\
& Strong Majority (4/5) & 196 & 155 & 0 & 41 & 79.08 \\
& Unanimous (5/5) & 311 & 291 & 0 & 20 & 93.57 \\
\midrule
\multirow{6}{*}{\rotatebox[origin=c]{90}{%
\begin{tabular}{c}\\
\scriptsize Mixtral-8x7B\\
\end{tabular}}}
& Baseline (No majority) & 1000 & 695 & 0 & 305 & 69.50 \\
\cmidrule(lr){2-7}
& Minimal Majority (2/5) & 58 & 26 & 0 & 32 & 44.83 \\
& Moderate Majority (3/5) & 135 & 90 & 0 & 45 & 66.67 \\
& Other & 35 & 3 & 0 & 32 & 8.57 \\
& Split Vote (2/2) & 27 & 14 & 0 & 13 & 51.85 \\
& Strong Majority (4/5) & 175 & 142 & 0 & 33 & 81.14 \\
& Unanimous (5/5) & 470 & 437 & 0 & 33 & 92.98 \\
\midrule
\multirow{6}{*}{\rotatebox[origin=c]{90}{%
\begin{tabular}{c}
\\
\scriptsize gpt-oss-20B\\
\end{tabular}}}
& Baseline (No majority) & 1000 & 735 & 0 & 265 & 73.50 \\
\cmidrule(lr){2-7}
& Minimal Majority (2/5) & 50 & 25 & 0 & 25 & 50.00 \\
& Moderate Majority (3/5) & 110 & 79 & 0 & 31 & 71.82 \\
& Other & 28 & 4 & 0 & 24 & 14.29 \\
& Split Vote (2/2) & 20 & 11 & 0 & 9 & 55.00 \\
& Strong Majority (4/5) & 160 & 135 & 0 & 25 & 84.38 \\
& Unanimous (5/5) & 532 & 506 & 0 & 26 & 95.11 \\
\midrule
\multirow{6}{*}{\rotatebox[origin=c]{90}{%
\begin{tabular}{c}
\\
\scriptsize DeepSeek-Qwen\\
\end{tabular}}}
& Baseline (No majority) & 1000 & 496 & 0 & 304 & 62.00 \\
\cmidrule(lr){2-7}
& Minimal Majority (2/5) & 142 & 37 & 0 & 105 & 26.06 \\
& Moderate Majority (3/5) & 125 & 86 & 0 & 39 & 68.80 \\
& Other & 66 & 13 & 0 & 53 & 19.70 \\
& Split Vote (2/2) & 43 & 10 & 0 & 33 & 23.26 \\
& Strong Majority (4/5) & 110 & 92 & 0 & 18 & 83.64 \\
& Unanimous (5/5) & 314 & 299 & 0 & 15 & 95.22 \\
\bottomrule
\end{tabular}}
\caption{GSM8K Results.}
\label{tab:gsm8k}
\end{table}

\begin{table}[ht]
\centering
\footnotesize
\setlength{\tabcolsep}{3pt}
\resizebox{\textwidth}{!}{%
\begin{tabular}{@{}lcccccc@{}}
\toprule
& & Question Count & Correct & No Answer & Incorrect & Accuracy (\%) \\
\midrule
\multirow{6}{*}{\rotatebox[origin=c]{90}{%
\begin{tabular}{c}
\\
\scriptsize Qwen2.5-7B\\
\end{tabular}}}
& Baseline (No majority) & 1000 & 666 & 0 & 234 & 74.00 \\
\cmidrule(lr){2-7}
& Minimal Majority (2/5) & 82 & 41 & 0 & 41 & 50.00 \\
& Moderate Majority (3/5) & 141 & 94 & 0 & 47 & 66.67 \\
& Other & 40 & 3 & 0 & 37 & 7.50 \\
& Split Vote (2/2) & 31 & 10 & 0 & 21 & 32.26 \\
& Strong Majority (4/5) & 170 & 155 & 0 & 15 & 91.18 \\
& Unanimous (5/5) & 436 & 421 & 0 & 15 & 96.56 \\
\midrule
\multirow{6}{*}{\rotatebox[origin=c]{90}{%
\begin{tabular}{c}
\\
\scriptsize Llama-3.1-8B\\
\end{tabular}}}
& Baseline (No majority) & 1000 & 657 & 0 & 343 & 65.70 \\
\cmidrule(lr){2-7}
& Minimal Majority (2/5) & 104 & 64 & 0 & 40 & 61.54 \\
& Moderate Majority (3/5) & 237 & 155 & 0 & 82 & 65.40 \\
& Other & 41 & 10 & 0 & 31 & 24.39 \\
& Split Vote (2/2) & 42 & 13 & 0 & 29 & 30.95 \\
& Strong Majority (4/5) & 273 & 244 & 0 & 29 & 89.38 \\
& Unanimous (5/5) & 303 & 279 & 0 & 24 & 92.08 \\
\midrule
\multirow{6}{*}{\rotatebox[origin=c]{90}{%
\begin{tabular}{c}
\\
\scriptsize Mixtral-8x7B\\
\end{tabular}}}
& Baseline (No majority) & 1000 & 720 & 0 & 280 & 72.00 \\
\cmidrule(lr){2-7}
& Minimal Majority (2/5) & 70 & 38 & 0 & 32 & 54.29 \\
& Moderate Majority (3/5) & 165 & 115 & 0 & 50 & 69.70 \\
& Other & 42 & 5 & 0 & 37 & 11.90 \\
& Split Vote (2/2) & 33 & 12 & 0 & 21 & 36.36 \\
& Strong Majority (4/5) & 210 & 188 & 0 & 22 & 89.52 \\
& Unanimous (5/5) & 380 & 362 & 0 & 18 & 95.26 \\
\midrule
\multirow{6}{*}{\rotatebox[origin=c]{90}{%
\begin{tabular}{c}
\\
\scriptsize gpt-oss-20B\\
\end{tabular}}}
& Baseline (No majority) & 1000 & 760 & 0 & 240 & 76.00 \\
\cmidrule(lr){2-7}
& Minimal Majority (2/5) & 60 & 35 & 0 & 25 & 58.33 \\
& Moderate Majority (3/5) & 130 & 98 & 0 & 32 & 75.38 \\
& Other & 35 & 7 & 0 & 28 & 20.00 \\
& Split Vote (2/2) & 25 & 12 & 0 & 13 & 48.00 \\
& Strong Majority (4/5) & 190 & 175 & 0 & 15 & 92.11 \\
& Unanimous (5/5) & 460 & 444 & 0 & 16 & 96.52 \\
\midrule
\multirow{6}{*}{\rotatebox[origin=c]{90}{%
\begin{tabular}{c}
\\
\scriptsize DeepSeek-Qwen\\
\end{tabular}}}
& Baseline (No majority) & 1000 & 645 & 0 & 355 & 64.50 \\
\cmidrule(lr){2-7}
& Minimal Majority (2/5) & 155 & 58 & 0 & 97 & 37.42 \\
& Moderate Majority (3/5) & 145 & 102 & 0 & 43 & 70.34 \\
& Other & 72 & 18 & 0 & 54 & 25.00 \\
& Split Vote (2/2) & 48 & 14 & 0 & 34 & 29.17 \\
& Strong Majority (4/5) & 135 & 122 & 0 & 13 & 90.37 \\
& Unanimous (5/5) & 345 & 333 & 0 & 12 & 96.52 \\
\bottomrule
\end{tabular}}
\caption{GSM-Symbolic Results.}
\label{tab:gsm_symbolic}
\end{table}

\begin{table}[ht]
\centering
\footnotesize
\setlength{\tabcolsep}{3pt}
\resizebox{\textwidth}{!}{%
\begin{tabular}{@{}lcccccc@{}}
\toprule
& & Question Count & Correct & No Answer & Incorrect & Accuracy (\%) \\
\midrule
\multirow{6}{*}{\rotatebox[origin=c]{90}{%
\begin{tabular}{c}
\\
\scriptsize Qwen2.5-7B\\
\end{tabular}}}
& Baseline (No majority) & 1000 & 586 & 0 & 314 & 65.11 \\
\cmidrule(lr){2-7}
& Minimal Majority (2/5) & 12 & 2 & 0 & 10 & 16.67 \\
& Moderate Majority (3/5) & 181 & 94 & 0 & 87 & 51.93 \\
& Split Vote (2/2) & 47 & 10 & 0 & 37 & 21.28 \\
& Strong Majority (4/5) & 205 & 132 & 0 & 73 & 64.39 \\
& Unanimous (5/5) & 455 & 397 & 0 & 58 & 87.25 \\
\midrule
\multirow{6}{*}{\rotatebox[origin=c]{90}{%
\begin{tabular}{c}
\\
\scriptsize Llama-3.1-8B\\
\end{tabular}}}
& Baseline (No majority) & 1000 & 483 & 0 & 417 & 53.67 \\
\cmidrule(lr){2-7}
& Minimal Majority (2/5) & 41 & 16 & 0 & 25 & 39.02 \\
& Moderate Majority (3/5) & 285 & 141 & 0 & 144 & 49.47 \\
& Other & 1 & 0 & 0 & 1 & 0.00 \\
& Split Vote (2/2) & 102 & 29 & 0 & 73 & 28.43 \\
& Strong Majority (4/5) & 228 & 173 & 0 & 55 & 75.88 \\
& Unanimous (5/5) & 243 & 208 & 0 & 35 & 85.60 \\
\midrule
\multirow{6}{*}{\rotatebox[origin=c]{90}{%
\begin{tabular}{c}
\\
\scriptsize Mixtral-8x7B\\
\end{tabular}}}
& Baseline (No majority) & 900 & 429 & 0 & 471 & 47.67 \\
\cmidrule(lr){2-7}
& Minimal Majority (2/5) & 27 & 12 & 0 & 15 & 44.44 \\
& Moderate Majority (3/5) & 291 & 128 & 0 & 163 & 43.99 \\
& Split Vote (2/2) & 105 & 27 & 0 & 78 & 25.71 \\
& Strong Majority (4/5) & 246 & 145 & 0 & 101 & 58.94 \\
& Unanimous (5/5) & 231 & 179 & 0 & 52 & 77.49 \\
\midrule
\multirow{6}{*}{\rotatebox[origin=c]{90}{%
\begin{tabular}{c}
\\
\scriptsize gpt-oss-20B\\
\end{tabular}}}
& Baseline (No majority) & 1000 & 368 & 0 & 632 & 36.80 \\
\cmidrule(lr){2-7}
& Minimal Majority (2/5) & 12 & 4 & 0 & 8 & 33.33 \\
& Moderate Majority (3/5) & 180 & 65 & 0 & 115 & 36.11 \\
& Other & 6 & 2 & 0 & 4 & 33.33 \\
& Split Vote (2/2) & 52 & 19 & 0 & 33 & 36.54 \\
& Strong Majority (4/5) & 220 & 82 & 0 & 138 & 37.27 \\
& Unanimous (5/5) & 430 & 168 & 0 & 262 & 39.07 \\
\midrule
\multirow{6}{*}{\rotatebox[origin=c]{90}{%
\begin{tabular}{c}
\\
\scriptsize DeepSeek-Qwen\\
\end{tabular}}}
& Baseline (No majority) & 1000 & 487 & 0 & 413 & 54.11 \\
\cmidrule(lr){2-7}
& Minimal Majority (2/5) & 27 & 8 & 0 & 19 & 29.63 \\
& Moderate Majority (3/5) & 246 & 99 & 0 & 147 & 40.24 \\
& Other & 4 & 0 & 0 & 4 & 0.00 \\
& Split Vote (2/2) & 76 & 35 & 0 & 41 & 46.05 \\
& Strong Majority (4/5) & 226 & 128 & 0 & 98 & 56.64 \\
& Unanimous (5/5) & 321 & 268 & 0 & 53 & 83.49 \\
\bottomrule
\end{tabular}}
\caption{MMLU Results}
\label{tab:mmlu}
\end{table}

\FloatBarrier

\section{Ablations}
\label{sec:ablation}

To validate our key hyperparameter choices, we conducted ablation studies on the two parameters that fundamentally define Noisy Coconut: the noise scale ($\sigma_0$) and the number of reasoning paths ($K$). Experiments were conducted on GSM8K using Qwen2.5-7B-Instruct. Table~\ref{tab:core_ablation} summarizes our findings.

\begin{table}[H]
\centering
\caption{Core ablation study on GSM8K (Qwen2.5-7B-Instruct) where number of questions is equal to 1000. We vary noise scale and path count independently. Baseline uses $K$=1 with no noise injection.}
\label{tab:core_ablation}
\begin{tabular}{lcc}
\toprule
\textbf{Configuration} & \textbf{Accuracy (\%)} & \textbf{$\Delta$} \\
\midrule
Baseline ($K$=1, $\sigma_0$=0) & 71.0 & --- \\
\midrule
\multicolumn{3}{l}{\textit{Noise scale (fixed $K$=5):}} \\
\quad $\sigma_0$=0.0 & 76.4 & +5.4 \\
\quad $\sigma_0$=0.2 & 78.2 & +7.2 \\
\quad $\sigma_0$=0.5 & 77.8 & +6.8 \\
\midrule
\multicolumn{3}{l}{\textit{Path count (fixed $\sigma_0$=0.2):}} \\
\quad $K$=1 & 72.0 & +1.0 \\
\quad $K$=5 ($\geq$3/5) & 78.2 & +7.2 \\
\quad $K$=10 ($\geq$6/10) & 80.2 & +9.2 \\\\
\bottomrule
\end{tabular}
\end{table}

\textbf{Noise scale.} We evaluated $\sigma_0 \in \{0, 0.2, 0.5\}$ with $K$=5 paths. Setting $\sigma_0$=0 (i.e., multiple paths without noise injection) yields 76.4\% accuracy (+5.4\% over baseline), demonstrating that diversity alone provides substantial improvements through majority voting. However, performance peaks at $\sigma_0$=0.2 with 78.2\% accuracy, indicating that noise injection provides an additional +1.8\% gain beyond path diversity alone. Higher noise scales show logarithmic decay, with $\sigma_0$=0.5 achieving 77.8\% as individual path quality degrades. We select $\sigma_0$=0.2 as it achieves the optimal balance between exploration diversity and reasoning coherence.

\textbf{Number of paths.} We evaluated $K \in \{1, 5, 10\}$ with $\sigma_0$=0.2. Performance scales sublinearly with $K$, showing diminishing returns beyond $K$=5. While $K$=10 achieves 2.0 points higher accuracy, it requires 2$\times$ the computational cost. We select $K$=5 as a practical tradeoff between performance and efficiency.

\FloatBarrier

\clearpage
\section{Language Mixing}
\label{sec:language_mixing}

\begin{questionbox}
\textbf{Question ID:} 1 \hfill \textbf{Reference Answer:} 285\\[0.5em]
Gabriel is popping popcorn for a snack. As the skillet of kernels heats up, the kernels start popping faster. 30 pop in the first 20 seconds of cooking, then 2 times that amount in the next 20 seconds. The kernels increase to 4 times the initial popping rate in the next 20 seconds, but in the final 20 seconds, the popping slows down to half the rate as the past 20 seconds. After Gabriel takes the skillet off the heat, a quarter of the number of kernels that popped in the final 20 seconds of cooking also pop from the residual heat. How many pieces of popcorn does Gabriel have to eat?
\end{questionbox}

\begin{englishbox}
Initial popping rate is 30 kernels in 20 seconds.

In the next 20 seconds, the popping rate is 2 times the initial rate:
\[2 \times 30 = 60 \text{ kernels}\]

In the next 20 seconds, the popping rate is 4 times the initial rate:
\[4 \times 30 = 120 \text{ kernels}\]

In the final 20 seconds, the popping rate is half of the previous 20 seconds:
\[\frac{1}{2} \times 60 = 30 \text{ kernels}\]
\textcolor{red}{\textit{$\leftarrow$ Error: Should be half of 120, not 60}}

After taking the skillet off the heat, a quarter of the final 20 seconds' popping also pops from the residual heat:
\[\frac{1}{4} \times 30 = 7.5 \text{ kernels}\]

Adding all the popped kernels together:
\[30 + 60 + 120 + 30 + 7.5 = 247.5 \text{ kernels}\]

Since we can't have half a kernel, we'll round down to the nearest whole number:

\begin{resultbox}[incorrectbg]
\centering\textbf{Final Answer: 247} \quad \textcolor{red}{\ding{55} Incorrect}
\end{resultbox}
\end{englishbox}

\noindent\includegraphics[width=\textwidth]{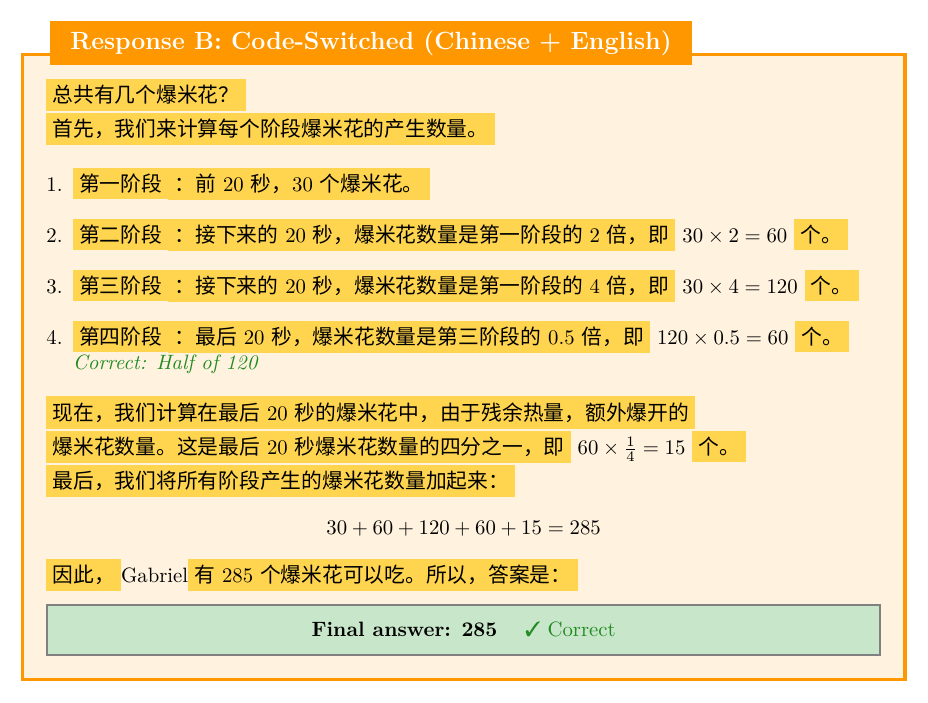}

\begin{translationbox}
\textit{How many pieces of popcorn are there?}

\textit{First, let's calculate the popping rate for each step:}

\begin{enumerate}[leftmargin=*]
    \item \textit{\textbf{First step}: the first 20 seconds, 30 pieces of popcorn.}
    \item \textit{\textbf{Second step}: In the next 20 seconds, the number of popped kernels is 2 times the amount from the first step,} i.e.\ $30 \times 2 = 60$ \textit{kernels.}
    \item \textit{\textbf{Third step}: In the next 20 seconds, the number of popped kernels is 4 times the amount from the first step,} i.e.\ $30 \times 4 = 120$ \textit{kernels.}
    \item \textit{\textbf{Fourth step}: In the last 20 seconds, the number of popped kernels is 0.5 times the amount from the third step,} i.e.\ $120 \times 0.5 = 60$ \textit{kernels.}
\end{enumerate}

\textit{Now, we calculate the number of popped kernels in the last 20 seconds' popping from the residual heat. This is a quarter of the final 20 seconds' popping, meaning that} $60 \times \frac{1}{4} = 15$ \textit{kernels.}

\textit{Finally, we sum up all the numbers of popped kernels from every step:}
\[30 + 60 + 120 + 60 + 15 = 285\]

\textit{Thus, Gabriel has 285 pieces of popcorn that he can eat. Therefore, the answer is:}

\begin{resultbox}[correctbg]
\centering\textbf{Final answer: 285} \quad \textcolor{ForestGreen}{\ding{51} Correct}
\end{resultbox}
\end{translationbox}

\begin{center}
\begin{tcolorbox}[
    enhanced,
    width=0.9\textwidth,
    colback=gray!10,
    colframe=gray!50,
    sharp corners,
    boxrule=0.5pt
]
\centering
\small\textit{Figure: Comparison of model outputs demonstrating noise-induced code-switching behavior. At certain noise thresholds, the model exhibits language mixing (Chinese-English) which correlates with improved reasoning accuracy on this mathematical task. Yellow highlighting indicates Chinese text segments in the code-switched response. Translation provided by Hannah Cyberey.}
\end{tcolorbox}
\end{center}

\end{document}